\PassOptionsToPackage{dvipsnames}{xcolor}
\documentclass[10pt,logo,copyright]{nvidiatechreport}

\usepackage{amsmath,amsfonts,bm}
\usepackage{tikz} 

\def\eqref#1{equation~\ref{#1}}

\def\1{\bm{1}}

\DeclareMathAlphabet{\mathsfit}{\encodingdefault}{\sfdefault}{m}{sl}
\SetMathAlphabet{\mathsfit}{bold}{\encodingdefault}{\sfdefault}{bx}{n}

\usepackage{natbib}

\usepackage{enumitem}

\usepackage{wrapfig}
\usepackage{booktabs}
\usepackage{xspace}
\usepackage{multirow}
\usepackage{xcolor}
\usepackage{xspace}
\usepackage{comment}
\usepackage{url}
\usepackage{tikz}
\usetikzlibrary{positioning,fit,backgrounds,arrows.meta,calc,patterns,matrix}
\usepackage{float}
\usepackage[most]{tcolorbox}
\usepackage{multirow}   
\usepackage{graphicx}   

\usepackage{tabularx}
\usepackage{adjustbox}
\usepackage{makecell}
\usepackage{siunitx}
\usepackage{array}
\usepackage{threeparttable}
\usepackage[table]{xcolor}   
\usepackage[table]{xcolor} 
\usepackage{xcolor,colortbl}

\title{Front-Loading Reasoning: The Synergy between Pretraining and Post-Training Data}

\author{
Syeda Nahida Akter$^{2}$\thanks{Work done during internship at NVIDIA},~~ Shrimai Prabhumoye$^{1,3}$,~~ Eric Nyberg$^{2}$,~~ Mostofa Patwary$^{1}$,\\
\textbf{Mohammad Shoeybi$^{1}$,~~ Yejin Choi$^{1,4}$,~~ Bryan Catanzaro$^{1}$}\\
NVIDIA$^{1}$, Carnegie Mellon University$^{2}$, Boston University$^{3}$, Stanford University$^{4}$\\
\texttt{sakter@andrew.cmu.edu},~~\texttt{sprabhumoye@nvidia.com}
}

\newtcolorbox{takeaway}[3][]{
  enhanced,
  sharp corners,
  boxrule=0pt,
  colback=#2!4!white,
  borderline west={3.5pt}{0pt}{#2!85!black},
  drop fuzzy shadow=black!30!white,
  arc=2.6mm,
  left=6mm, right=6mm, top=4mm, bottom=4mm,
  coltitle=black,
  title={\small\bfseries #3}, 
  attach boxed title to top left={yshift=-3.0mm, xshift=4.0mm}, 
  boxed title style={
    boxrule=0.4pt,
    colframe=#2!70!black,
    colback=#2!10!white, 
    rounded corners,
    drop shadow=black!15,
    top=0.6mm, bottom=0.6mm, left=1.8mm, right=1.8mm 
  },
  before skip=2mm, after skip=2mm,
  #1
}

\begin{document}

\maketitle

\newcommand{\todo}[1]{{\color{red}\bf [TODO: #1]}\xspace}
\newcommand{\shrimai}[1]{{\color{red}\bf [Shrimai: #1]}\xspace}

\newcommand{\syeda}[1]{{\color{blue}\bf [Syeda: #1]}\xspace}

\newcommand{\shrimaix}[1]
{\draftcomment{\textcolor{magenta}{Shrimai: \sout{#1}}}}

\newcommand{\syedax}[1]
{\draftcomment{\textcolor{gray}{Syeda: \sout{#1}}}}

\newcommand{\yejin}[1]{{\color{cyan} [#1]$_{yejin}$}}

\newcommand{\ptgpr}{\textsc{gpr}$_{\mathtt{PT}}$ \textsc{avg}\xspace}
\newcommand{\ptmath}{\textsc{math}$_{\mathtt{PT}}$ \textsc{avg}\xspace}
\newcommand{\ptcode}{\textsc{code}$_{\mathtt{PT}}$ \textsc{avg}\xspace}
\newcommand{\ptsci}{\textsc{science}$_{\mathtt{PT}}$ \textsc{avg}\xspace}
\newcommand{\sftgpr}{\textsc{gpr}$_{\mathtt{SFT}}$ \textsc{avg}\xspace}
\newcommand{\sftmath}{\textsc{math}$_{\mathtt{SFT}}$ \textsc{avg}\xspace}
\newcommand{\sftcode}{\textsc{code}$_{\mathtt{SFT}}$ \textsc{avg}\xspace}
\newcommand{\sftsci}{\textsc{science}$_{\mathtt{SFT}}$ \textsc{avg}\xspace}
\newcommand{\sftins}{\textsc{ins}$_{\mathtt{SFT}}$ \textsc{avg}\xspace}
\newcommand{\vlm}{\textsc{vlm}\xspace}
\newcommand{\grpo}{\textsc{grpo}\xspace}
\newcommand{\vlms}{\textsc{vlms}\xspace}
\newcommand{\llm}{\textsc{llm}\xspace}
\newcommand{\lmm}{\textsc{lmm}\xspace}
\newcommand{\llava}{\textsc{llava-1.5}\xspace}
\newcommand{\nc}{\textsc{nemotron-crossthink}\xspace}

\newcommand{\llama}{\textsc{LLaMA3-70B-Instruct}\xspace}
\newcommand{\llamas}{\textsc{LLaMA3-8B-Instruct}\xspace}
\newcommand{\llamabase}{\textsc{LLaMA3-70B}\xspace}

\newcommand{\sft}{\mathrm{SFT}\xspace}
\newcommand{\dres}{\mathrm{res}}
\newcommand{\dpt}{\mathrm{PT}}
\newcommand{\dsft}{\mathrm{SFT}}
\newcommand{\datad}{\mathcal{D}}
\newcommand{\thetafinal}{\mathrm{final}}
\newcommand{\dbase}{\mathrm{base}}
\newcommand{\dldq}{\mathrm{LDQ}}
\newcommand{\dshq}{\mathrm{SHQ}}
\newcommand{\dlmq}{\mathrm{LMQ}}
\newcommand{\dalf}{\mathrm{ALF}}

\newcommand{\owm}{OpenWebMath\xspace}
\newcommand{\mathpile}{\textsc{MathPile}\xspace}
\newcommand{\dsm}{\textsc{DeepSeekMath}\xspace}
\newcommand{\owma}{\textsc{owm}\xspace}
\newcommand{\mmlu}{\textsc{mmlu}\xspace}
\newcommand{\mmlus}{\textsc{mmlu-stem}\xspace}
\newcommand{\gpt}{\textsc{gpt-4v}\xspace}
\newcommand{\gemini}{\textsc{gemini pro}\xspace}
\newcommand{\cc}{CommonCrawl\xspace}
\newcommand{\gptllm}{\textsc{gpt-4}\xspace}
\newcommand{\gptllmold}{\textsc{gpt-3.5}\xspace}

\newcommand{\gsm}{\textsc{gsm8k}\xspace}

\begin{abstract}
The prevailing paradigm for enhancing the reasoning abilities of Large Language Models (LLMs) revolves around post-training on high-quality, reasoning-intensive data. While emerging literature suggests that reasoning data is increasingly incorporated also during the mid-training stage---a practice that is relatively more proprietary and less openly characterized---the role of such data in pretraining remains unclear. In particular, due to the opaqueness of pretraining corpora in most frontier models, the effect of reasoning data introduced at different phases of pre- and/or post-training is relatively less reported in the scientific literature. This raises several important but unsettled questions: \textit{Is adding reasoning data earlier during pre-training any better than introducing it during post-training, when the token counts are controlled? Could earlier inclusion risk overfitting and harm generalization, or instead establish durable foundations that later fine-tuning cannot recover?}  To address these questions, we conduct the first systematic study of \emph{how} reasoning data—varying in scale, diversity, and quality—affects LLM performance \emph{when} introduced at different stages of training. Our findings reveal that \textit{front-loading reasoning data into pretraining is critical (19\% average gain)}, establishing foundational capabilities that cannot be fully replicated by later-stage SFT, even with more data. We uncover an asymmetric principle for optimal data allocation: \textit{pretraining benefits most from broad diversity in reasoning patterns (11\% average gain), while SFT is more sensitive to data quality (15\% average gain with high quality data)}. Furthermore, we show that \textit{high-quality pretraining data has latent effects, activated only after SFT}, and that \textit{naively scaling SFT data can be detrimental}, washing away the benefits of early reasoning injection. Collectively, our results challenge the conventional separation of language modeling and reasoning, providing a principled guide for strategically allocating data across the entire training pipeline to build more capable models.

\end{abstract}
\abscontent

\section{Introduction}



The reasoning abilities of Large Language Models (LLMs) have advanced considerably, with post-training on reasoning data driving significant breakthroughs in reasoning tasks, such as math competitions \citep{hendrycksmath2021}, PhD-level scientific QA \citep{rein2024gpqa, phan2025humanitysexam}, and software engineering \citep{jimenez2024swebench}. 
This progress has been largely driven by mid- or post-training LLMs on high-quality, reasoning-intensive datasets---often featuring long chain-of-thought (CoT) examples \citep{guha2025openthoughtsdatarecipesreasoning,moshkov2025aimo2winningsolutionbuilding,zhou2025megamathpushinglimitsopen, gandhi2025cognitive, wang2025octothinker}. 
While this approach has proven effective, it treats reasoning as a specialized skill to be layered onto a generalist base. 
In addition, the impact of incorporating reasoning data during pretraining---and the potential synergistic effects on subsequent post-training---remains a critical yet less explored frontier. This research gap persists due to the prohibitive computational cost of end-to-end pretraining experiments and the opacity surrounding proprietary training recipes, which has concentrated community efforts on the more accessible post-training phase. 

The synergy between post-training phases has been widely explored \citep{liu2025acereasonnemotron11advancingmath, chen2025synergydilemmalongcotsft, chu2025sftmemorizesrlgeneralizes}, yet conclusions vary with training data and scale, and their applicability to pretraining remains vague in the current literature.
In this work, we investigate not just \emph{which} reasoning data, but \emph{when} to train with such reasoning data
by studying the synergy 
between pretraining and post-training. 
Our central goal is to determine the ideal balance of such reasoning data across the two phases in order to maximize downstream accuracies after reinforcement learning. This motivates the following research questions: 


\begin{itemize}[leftmargin=*]
    \item\textit{Is a reasoning-rich pretraining essential, or can a model ``catch up"?} We investigate whether a model pretrained without reasoning data can match the performance of its reasoning-aware counterparts by simply undergoing a more intensive SFT phase.
    \item \textit{Does inclusion of reasoning data make the base \llm overfitted and less generalizable to sustain gains in subsequent training phases?} While recent literature highlights overspecialization of reasoning during post-training can be detrimental~\citep{gupta2025selectiveselftosupervisedfinetuninggeneralization, luo2025empiricalstudycatastrophicforgetting}, investigations of this effect in pretraining remain limited.
    \item \textit{Does data diversity in pretraining impact stability and specialization during SFT?} Specifically, does using the same reasoning data in both pretraining and SFT lead to robust skill mastery, or does a \textit{narrow pretraining focus} risk catastrophic forgetting when the model is later fine-tuned on different tasks?
    \item \textit{Does the complexity and quality of reasoning data matter when incorporated during pretraining of the base model?} Current literature explores this mostly from SFT stage \citep{zhou2023lima,guha2025openthoughtsdatarecipesreasoning}, making it obscure whether difficulty or noisiness in the early phase of training directly impacts reasoning development or not.
\end{itemize}


This work provides a systematic analysis of the interplay between reasoning data and the distinct phases of LLM training. Our primary findings are summarized as:

\begin{figure}[t]
  \centering
  \includegraphics[width=0.7\columnwidth]{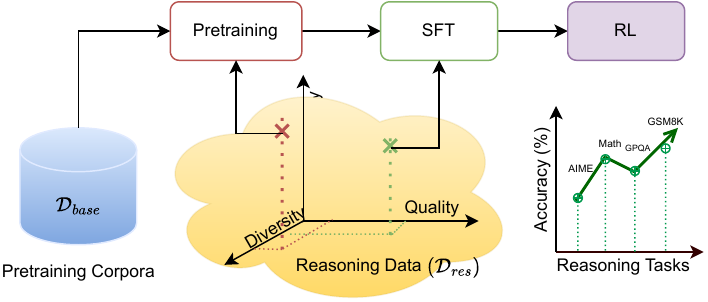}
  \caption{We systematically inject reasoning-style data ($\datad_{\dres}$) at different phases of training—pretraining versus $\sft$—while varying its \emph{diversity, quantity, and quality}. Our results show an asymmetric principle: diversity and scale matter most during pretraining, whereas quality dominates in SFT. This allocation strategy compounds through reinforcement learning (RL), yielding sustained gains across complex reasoning benchmarks.}
  \vspace{-1em}
  \label{fig:method}
\end{figure}

\begin{itemize}[leftmargin=*]
    \item \textbf{Front-loading reasoning data into pretraining creates a durable, compounding advantage.} Injecting reasoning data during pretraining establishes a superior foundation that widens at every stage of post-training, culminating in a \textbf{+19\%} lead on expert-level benchmarks. This refutes the \textit{catch-up} and \textit{overfitting} hypotheses, proving that SFT cannot compensate for a weak foundation and that pretraining choices dictate the final performance ceiling.

    \item \textbf{The optimal data strategy is asymmetric: prioritize diversity in pretraining and quality in SFT.} Our results reveal a clear, phase-dependent principle. Pretraining benefits most from \textbf{diversity and scale} (a \textbf{+11\%} gain with diverse corpus), while SFT is dominated by \textbf{data quality} (a \textbf{+15\%} gain with high-quality reasoning data). This provides an actionable heuristic for data allocation that is more nuanced than simplistic ``more is better'' approaches.

    \item \textbf{Naive scaling of SFT data is ineffective and harmful.} Blindly scaling SFT with mixed-quality data 
    yields no average improvement and actively harmed mathematical reasoning by \textbf{-5\%} on average, while a marginal (0.4\%) addition of high-quality data consistently improved performance. 

    \item \textbf{High-quality pretraining data can have a latent effect unlocked by SFT.} We found that high-quality data added to a diverse pretraining mix showed minimal immediate benefit but ``unlocked'' an additional \textbf{+4\%} gain over model pretrained with diverse, mixed quality data after SFT---revealing a deeper synergy where pretraining can instill a latent potential in the model that is only activated during the alignment phase.
\end{itemize}

\section{Methodology}
\label{sec:method}

Our methodology is designed to systematically determine the optimal strategy for allocating reasoning data between the pretraining and supervised fine-tuning stages of \llm development. 
We frame this as an optimization problem where the goal is to maximize the final model's downstream accuracies, $\mathcal{P}$. This is a function of the reasoning data introduced during pretraining, $\datad_{\dres}^{\dpt}$, and the data used for supervised fine-tuning, $\datad_{\dres}^{\dsft}$. Our objective is to find the optimal data configurations, $(\datad_{\dres}^{\dpt*}, \datad_{\dres}^{\dsft*})$, that solves the following:
\[
(\datad_{\dres}^{\dpt*}, \datad_{\dres}^{\dsft*}) = \arg\max_{\datad_{\dres}^{\dpt}, \datad_{\dres}^{\dsft}} \mathcal{P}(\theta_{\thetafinal})
\]
where $\theta_{\thetafinal}$ represents the parameters of the final model trained on data recipes defined by the choice of reasoning data at both stages.

Let $\datad_{\dbase}$ denote the general pretraining corpus and we define a model $\mathcal{M}(\theta)$ with parameters $\theta$ trained in two stages:  


\vspace{-2em}
\begin{align*}
\textbf{Pretraining:} \quad 
& \theta_{\dpt} = \arg\min_{\theta} \; \mathbb{E}_{(x,y) \sim \datad_{\dbase} \cup \datad_{\dres}^{\dpt}}\; \mathcal{L}_{\text{LM}}(f_\theta(x), y), \\
\textbf{SFT:} \quad 
& \theta_{\dsft} = \arg\min_{\theta} \; \mathbb{E}_{(x,y) \sim \datad_{\dres}^{\dsft}} \; \mathcal{L}_{\dsft}(f_\theta(x), y),
\end{align*}

\paragraph{Evaluation Objective.}  
The central research question can be expressed as analyzing the function:
\begin{equation}
\mathcal{P}(\datad_{\dres}^{\dpt}, \datad_{\dres}^{\dsft}) 
= \mathbb{E}_{t \sim \mathcal{T}} \Big[ \text{Acc}\big(f_{\theta_{\dsft}}(t)\big) \Big],
\end{equation}
where $\mathcal{T}$ is a set of downstream reasoning tasks (math, science, code, general reasoning).

Our study can be summarized as optimizing the allocation of $\datad_{\dres}$ between pretraining and SFT:
\begin{equation}
\max_{\;\datad_{\dres}^{\dpt}, \; \datad_{\dres}^{\dsft}} 
\;\; \mathcal{P}(\datad_{\dres}^{\dpt}, \datad_{\dres}^{\dsft})
\quad \text{s.t.} \quad 
\mathcal{B} = |\datad_{\dres}^{\dpt}| + |\datad_{\dres}^{\dsft}|,
\end{equation}
where $\mathcal{B}$ is the total budget of reasoning data available.  

This captures the trade-off of early, scale/diversity vs late, quality/complexity: $ \datad_{\dres}^{\dpt} \longleftrightarrow \datad_{\dres}^{\dsft}$




\subsection{Model Architecture and Baseline}

We select a hybrid transformer with a mixture of Mamba 2 \citep{pmlr-v235-dao24a}, self-attention and FFN layers \citep{nvidia2025nemotronhfamilyaccurateefficient} with an 8B parameter for our base model, $\mathcal{M}$ and pretrain from scratch for 1 trillion tokens. This size strikes a balance between computational feasibility and the capacity to learn complex reasoning patterns. 

\subsection{Data Pipeline}
\label{ss:dp}

Our experimental design relies on a careful distinction between two categories of data: (1) a large-scale, general-purpose pretraining corpus, and (2) a reasoning-focused, instruction-tuning (SFT-style) datasets of varying quality and scale. This separation allows us to precisely control the injection of reasoning data at different stages of training.

\paragraph{General Pretraining Corpus ($\datad_{\dbase}$).} For the base training corpus, we adopt the dataset introduced in \citet{nvidia2025nvidianemotronnano2}, which contains 6.2T tokens drawn from high-quality Common Crawl, mathematics, and code sources. This corpus provides broad coverage of languages and technical domains, serving as the backbone of all pretraining experiments.

\paragraph{Reasoning Datasets ($\datad_{\dres}$).} 
To investigate the impact of data quality, diversity, and complexity, we curate four distinct reasoning-focused datasets in the question-answer format:

\begin{itemize}[leftmargin=*]
    \item \textbf{Large-Scale, Diverse Data ($\datad_{\dldq}$).}  
    To simulate a \textit{``quantity-over-quality"} strategy, we employ the Nemotron-Pretraining-SFT-v1 dataset \citep{nvidia2025nvidianemotronnano2}.
    This massive 336 billion token dataset offers extensive domain coverage, with a composition of approximately 56\% math, 17\% code, and 27\% science and general-purpose reasoning.
    The dataset covers tasks ranging from simple Q\&A to multi-turn dialogues, but with heterogeneous quality and reasoning depth, reflecting large-scale real-world availability.
    
    \item \textbf{Small-Scale, High-Quality Data ($\datad_{\dshq}$).}  
    To capture the effect of long chain-of-thought traces from strong teacher models, we include the dataset of \citet{guha2025openthoughtsdatarecipesreasoning}, comprising 1.2M carefully curated examples (71\% math, 21\% code, 8\% science). Compared to $\datad_{\dlmq}$, this corpus is smaller, less diverse, but significantly higher quality, emphasizing detailed reasoning paths.
    
    \item \textbf{Large-Scale, Mixed-Quality Data ($\datad_{\dlmq}$).}  
    To balance diversity with quality, we construct a combined dataset that is a straightforward union of the two datasets above: $\datad_{\dlmq} = \datad_{\dldq} \cup \datad_{\dshq}$. This mix preserves large-scale coverage while injecting a fraction of curated, high-quality reasoning traces.
    
    \item \textbf{Answer-Length Filtered Data ($\datad_{\dalf}$).}  
    To investigate the feature of data quality, we create a subset of $\mathbf{\mathcal{D}_{\text{LLQ}}}$ by retaining examples where the answer length exceeds 4096 tokens, based on the principle that longer responses often correspond to more complex CoT reasoning. This dataset allows us to isolate the impact of reasoning complexity in different training phases. 
\end{itemize}


\subsection{Synergy between Pretraining and SFT}

In this work, we aim to disentangle the contribution of reasoning data when incorporated at different points in the training pipeline. We structure the study into three stages: (i) large-scale \textbf{Pretraining}, where reasoning data may or may not be injected alongside the base corpus, (ii) \textbf{Supervised Finetuning (SFT)}, where pretrained models are further adapted on reasoning data of varying quality and diversity, and (iii) \textbf{Reinforcement Learning (RLVR)} to determine the sustainability of early reasoning gain in the final model. This subsection details the pretraining design; the SFT stage is described in the following section.

\paragraph{Phase 1: Pretraining.}

Prior work has primarily explored reasoning supervision either on top of fully pretrained {\llm}s \citep{wang2025octothinker} or by introducing small amounts of long chain-of-thought (CoT) data into intermediate checkpoints \citep{ai2025rethinkingreflectionpretraining}. These approaches leave open two questions: how to inject reasoning data at scale during end-to-end pretraining, and whether the benefits persist when combined with high-quality base corpora. 
To address these questions, we pretrain all models \emph{from scratch} for 1T tokens using a mix of 80\% of $\datad_{\dbase}$ and different types of $\datad_{\dres}$ for 20\%. 

Based on the reasoning data introduced, we train four distinct models:

\begin{itemize}[leftmargin=*]
    \item \textbf{$\mathcal{M}_{\dbase}$:} 
    This model serves as our \textbf{baseline} and is pretrained without any reasoning data. 
    \item \textbf{$\mathcal{M}_{\dldq}$:} 
    Pre-trained with large-scale, diverse $\datad_{\dldq}$ reasoning dataset along with $\datad_{\dbase}$. 
    \item \textbf{$\mathcal{M}_{\dshq}$:} 
    Pre-trained with $\datad_{\dshq}$ and $\datad_{\dbase}$ allowing us to isolate the effect of data quality versus the quantity and diversity of $\mathcal{M}_{\dldq}$.

    \item \textbf{$\mathcal{M}_{\dlmq}$:} 
    Finally, this model is exposed to our combined reasoning $\datad_{\dlmq}$ dataset.
\end{itemize}
In the subsequent analysis, we use $\mathcal{M}_{\dres}$ to denote the aggregate performance of the models pretrained with reasoning data, representing the average score across $\mathcal{M}_{\dshq}$, $\mathcal{M}_{\dldq}$, and $\mathcal{M}_{\dlmq}$.

\paragraph{Phase 2: Supervised Finetuning.}

Following pretraining, each of the four model variants ($\mathcal{M}_{\dbase}$, $\mathcal{M}_{\dldq}$, $\mathcal{M}_{\dshq}$, $\mathcal{M}_{\dlmq}$) is adapted through supervised finetuning (SFT). This second phase is crucial for understanding the synergies, redundancies, and trade-offs between the data introduced during pretraining versus the SFT stage. 
To this end, we design a controlled set of SFT experiments, where each pretrained model is finetuned on different reasoning corpora introduced in Section~\ref{ss:dp} to address the following rearch questions:

\begin{itemize}[leftmargin=*]
    \item \textbf{The ``Catch-Up'' Hypothesis:} Can intensive SFT on high-quality reasoning data allow the baseline model, $\mathbf{\mathcal{M}_{\text{base}}}$, to match or exceed the accuracy of models that were exposed to reasoning data during pretraining? This directly tests the criticality of early data injection versus late-stage specialization.
    
    \item \textbf{Impact of Pretraining Data Scale and Diversity:} We investigate how the scale and diversity of reasoning data used during pretraining influence the final model's capacity to absorb high-quality instruction data. Specifically, we ask: \textit{Does scaling up diverse reasoning data in pretraining provide lasting benefits even after all models are finetuned on the same high-quality SFT corpus?}
    By fine-tuning both the model pretrained on large, diverse data ($\mathcal{M}_{\dldq}$) and on smaller, less diverse data ($\mathcal{M}_{\dshq}$) on the same high-quality SFT set, we can determine whether a broad or a deep initial exposure to reasoning yields a better foundation for downstream specialization.
    
    \item \textbf{Impact of SFT Data Quality and Complexity:} By fine-tuning all four base models on datasets of varying quality ($\datad_{\dldq}$ vs. $\datad_{\dshq}$) and complexity ($\datad_{\dalf}$), we can measure the marginal utility of data quality at the SFT stage as a function of the model's initial pretraining condition.
\end{itemize}

This design enables us to address three critical dimensions: (1) the \textbf{synergy} between pretraining and SFT data, (2) the \textbf{marginal gains} of increasing SFT data scale when reasoning was already introduced in pretraining, and (3) the \textbf{role of data complexity and diversity} in determining whether reasoning supervision should be injected early, late, or across both stages.
Together with the pretraining experiments, these SFT studies form a fully crossed setup, providing the first systematic assessment of how reasoning-style SFT data interacts with  pretraining to shape the reasoning abilities of large language models.

\paragraph{Phase 3: Reinforcement Learning.}
To further observe the impact of reasoning centric pretraining and heavy supervised finetuning, we deploy RL using Group Relative Policy Optimization (\grpo) \citep{shao2024deepseekmathpushinglimitsmathematical} with verifiable rewards on top of the base models. Here we use \nc \cite{akter2025nemotroncrossthinkscalingselflearningmath} which has shown to be effective to enhance reasoning across diverse domains. 












\section{Experimental Setup}
\label{sec:setup}

\subsection{Training}

\paragraph{Pretraining.} To prepare base models, we pretrain a 8B \llm on our pretraining data blend till 1T tokens using 512 H100 80GB SXM5 GPUs. During training, we use the AdamW optimizer \citep{loshchilov2018decoupled} with $\beta_1 =0.9$, $\beta_2=0.95$ and weight decay of 0.1. We use a 8-way tensor and pipeline parallelism to train the model. We set the maximum value of learning rate to $3e^{-4}$, minimum 
to $3e^{-6}$, and use a batch size of 6M tokens with a 8192 context length.

\paragraph{Post-Training.} 
After pretraining, each 8B \llm is finetuned on 4.8M reasoning samples from $\datad_{res}$. SFT uses AdamW with $(\beta_1,\beta_2)=(0.9,0.95)$, weight decay $0.01$, warmup ratio $0.05$, learning rate $5{\times}10^{-6}$, batch size $512$, and context length $32$k. We then apply \grpo via the veRL framework\footnote{\url{https://github.com/volcengine/verl}} for one epoch on \nc data with constant LR $1{\times}10^{-6}$, batch size $128$, PPO mini-batch $128$, and context length $8192$. Each step samples $128$ prompts with $8$ rollouts (temperature$=1.0$, top-$p=1.0$), and a KL penalty coefficient of $0.001$.    



\subsection{Evaluation Metrics}

We report average accuracies of all tasks under each of the following categories.

\paragraph{Base Model Evaluations.} We conduct a thorough benchmark assessment to evaluate the generalizability of the base models, using a series of datasets using LM Eval Harness \citep{eval-harness}.

\begin{itemize}[leftmargin=*]

\item \textbf{General Purpose Reasoning (\ptgpr).} 
We consider four standard commonsense and logical reasoning tasks in 0-shot: ARC challenge \citep{clark2018thinksolvedquestionanswering}, HellaSwag \citep{zellers2019hellaswag}, WinoGrande \citep{sakaguchi2021winogrande}, and reading comprehension task: RACE \citep{lai-etal-2017-race}.

\item \textbf{Math Reasoning (\ptmath).} We evaluate the math reasoning ability with two benchmarks--they 
encompass math challenges from elementary to college level complexity demanding qualitative reasoning (8-shot GSM8K \citep{cobbe2021trainingverifierssolvemath}, 4-shot MATH-500 \citep{hendrycksmath2021}). 

\item \textbf{Science Reasoning (\ptsci).} We evaluate on 5-shot MMLU \citep{hendryckstest2021} and MMLU-Pro \citep{NEURIPS2024_ad236edc} that spans multiple domains, from professional to academic, testing the model on specialized subjects. 

\item \textbf{Code Reasoning (\ptcode).} For code tasks (HumanEval \citep{chen2021evaluatinglargelanguagemodels}, MBPP \citep{50670}) we evaluate the EvalPlus variants along with the sanitization of generations \citep{NEURIPS2023_43e9d647}, in a 0-shot setup. We estimate avg@32, pass@1 from 32 generations per prompt.
\end{itemize}

\paragraph{SFT Model Evaluations.} To evaluate the reasoning ability of different SFT models, we focus on reasoning centric benchmarks unlike in base model evaluations, where mostly focus on the generalizability of the \llm. We conduct evaluations using NeMo-Skills\footnote{\url{https://github.com/NVIDIA/NeMo-Skills}}.
\begin{itemize}[leftmargin=*]
    \item \textbf{Math Reasoning (\sftmath).} In addition to the GSM8K and MATH-500, we evaluate the models on two more complex math tasks---AIME24 and AIME25 \citep{aime_1983_2024}.
    \item \textbf{Science Reasoning (\sftsci).} On top of MMLU and MMLU-Pro, we evaluate on graduate level QA task: GPQA-Diamond \citep{rein2024gpqa}. 
    \item \textbf{Code Reasoning (\sftcode).} We choose LiveCodeBench \citep{jain2025livecodebench} to test complex code reasoning ability.
    \item \textbf{Instruction Following (\sftins).} For broader evaluation on diverse capabilities, we use IFEval \citep{zeng2024llmbar}.
\end{itemize}
We report Pass@1 average of 16 runs for AIME-2024, AIME-2025 and average of 4 runs for MATH-500, GSM8K, MMLU, MMLU-Pro, GPQA-Diamond, LiveCodeBench and IFEval.

\paragraph{RL Model Evaluations.} In this phase, we evaluate the models on complex reasoning tasks such as AIME24,25, MATH-500, GSM8K, MMLU, MMLU-Pro, GPQA-Diamond, LiveCodeBench following the evaluation metric in \textsc{sft} phase.
\section{Experiments and Results}
\label{sec:results}


\paragraph{Immediate Foundational Gains from Reasoning Data in Pretraining.}
Table~\ref{tab:PT_gains} shows the average accuracies of our four model variants immediately after the 1T token pretraining phase. The results provide clear evidence that integrating reasoning-style corpora from the start builds a significantly more capable foundation. 
Every model exposed to reasoning data surpasses baseline $\mathcal{M}_{\dbase}$. The largest improvements come from models trained on large-scale, diverse data; $\mathcal{M}_{\dldq}$ achieves highest average, driven by a +28.4\% gain in mathematics and a +9\% gain in code over the baseline. Interestingly, the smaller, less diverse, high-quality dataset ($\mathcal{M}_{\dshq}$) provides a modest lift, suggesting that at this early stage, the scale and diversity of the reasoning data are more critical than its curated quality for establishing a broad and robust reasoning foundation.

\begin{table*}[h!]
\begin{center}
\resizebox{0.75\textwidth}{!}{%
\begin{tabular}{@{}lccccc@{}}
\toprule
Model &  Average &\ptmath & \ptsci & \ptcode & \ptgpr\\\midrule
$\mathcal{M}_{\dbase}$ & 52.70 &   47.17	&47.13	&40.89	&75.63\\\midrule
$\mathcal{M}_{\dshq}$ & 54.98 &	52.60	&46.90	&44.32	&76.09\\
$\mathcal{M}_{\dldq}$ & 64.09	& 75.56	& 54.38	& 49.94	& 76.48\\
$\mathcal{M}_{\dlmq}$ & 64.07 &	72.37	& 54.49 & 52.60	& 76.83\\\midrule
$\mathcal{M}_{\dres}$ & \textbf{61.05} & 66.84 &	51.92	& 48.95	& 76.46\\
\toprule
\end{tabular}
}%
\end{center}\vspace{-1em}
\caption{\textbf{Average Accuracies of base models trained without or with varying $\datad_{\dres}$.} Pretraining with diverse reasoning data yields immediate gains, with scale and diversity driving math and code improvements more than quality.}
\label{tab:PT_gains}
\end{table*}

\paragraph{Pretraining Advantage is Maintained and Amplified Post-SFT.}
We evaluate whether a strong SFT phase can close the accuracy gap established during pretraining with diverse reasoning data $\datad_{\dres}$. 
At the same time, we examine whether the inclusion of such data causes the model to overfit and reduce generalization, thereby diminishing subsequent post-training gains. 
The results in \autoref{tab:SFT_gains} indicate that the advantage gained during the pre-training phase not only persists but is amplified. The group of models pretrained with reasoning data ($\mathcal{M}_{\dres}+\mathrm{SFT}$) outperforms the baseline group ($\mathcal{M}_{\dbase}+\mathrm{SFT}$) by a significant 9.3\% on average. This result strongly refutes the ``catch-up" hypothesis, showing that SFT is not a substitute for a strong reasoning foundation built during pretraining. While recent works have found reasoning-centric post-training to be most effective on math domains, the improvement on science is minimal \citep{prabhakar2025omnisciencedomainspecializedllmscientific, deepscaler2025, huan2025doesmathreasoningimprove}. However, the accuracy disparity in our findings is most prominent in science domains, an area often overlooked in reasoning-focused post-training work. This suggests that pretraining with reasoning data does more than teach facts; it helps the model develop effective internal representations for abstract and logical structures to enhance problem solving ability across domains. It does not overfit the model rather infuses the critical thinking ability that comes into full potential after post-training (Appendix \ref{sec:data_redun}). Consequently, the model's capacity to absorb and leverage the SFT data is fundamentally enhanced, leading to greater learning efficiency and a higher performance ceiling. SFT acts as a powerful enhancer, but its ultimate effectiveness is constrained by the quality of the foundation established during pretraining.

\begin{table*}[h!]
\begin{center}
\resizebox{0.85\textwidth}{!}{%
\begin{tabular}{@{}lccccc@{}}
\toprule
Model &  Average &\sftmath & \sftsci & \sftcode 	& \sftins\\\midrule
$\mathcal{M}_{\dbase} +\mathrm{SFT}$ & 26.62 & 34.48	& 20.92	& 7.09	& 43.98\\
$\mathcal{M}_{\dres}+\mathrm{SFT}$ & \textbf{35.92} & 40.61	&34.77	&16.75	&51.52\\
\toprule
\end{tabular}
}%
\end{center}
\vspace{-1em}
\caption{\textbf{Average Accuracies of SFT models pretrained with varying $\datad_{\dres}$.} SFT amplifies the pretraining advantage—models with reasoning-rich pretraining significantly outperform baseline.}
\label{tab:SFT_gains}
\end{table*}



\begin{table*}[h!]
\centering
\resizebox{\textwidth}{!}{%
\begin{tabular}{@{}lccccccccc@{}}
\toprule
\multirow{2}{*}{Model} & \multirow{2}{*}{Avg.} 
& \multicolumn{4}{c}{\textbf{Math Reasoning}} 
& \multicolumn{4}{c}{\textbf{Science \& Code Reasoning}} \\
\cmidrule(lr){3-6}\cmidrule(lr){7-10}
 &  & \textsc{math-500} & \textsc{gsm8k} & \textsc{aime24} & \textsc{aime25} 
 & \textsc{gpqa} & \textsc{mmlu} & \textsc{mmlu-pro} & \textsc{LCB} \\
\midrule
$\mathcal{M}_{\dbase}+\mathrm{SFT}+\mathrm{RL}$  
& 37.92 & 72.05 & 83.83 & 12.29 & 16.04 
& 28.16 & 41.10 & 36.69 & 13.16 \\
$\mathcal{M}_{\dlmq}+\mathrm{SFT}+\mathrm{RL}$ 
& \textbf{56.66} & 87.13 & 93.07 & 45.21 & 33.96
& 31.69 & 72.91 & 56.91 & 32.43 \\
\toprule
\end{tabular}%
}
\vspace{-1em}
\caption{\textbf{Average accuracies of RL models pretrained and fine-tuned with varying $\datad_{\dres}$.} 
Introducing reasoning data early provides significant reasoning boost after post-training.}
\label{tab:RL_gains}
\end{table*}

\paragraph{Pretraining Strategy Dictates Final Accuracy on Expert-Level Tasks.}
The final RL phase reveals the definitive impact of our pretraining interventions, particularly on expert-level reasoning benchmarks. 
We select $\mathcal{M}_{\dlmq}+\mathrm{SFT}$ and $\mathcal{M}_{\dbase}+\mathrm{SFT}$ using $\datad_{\dshq}$ as our two extreme pretraining backbones.
As shown in Table~\ref{tab:RL_gains}, the accuracy gap between the two models continues to diverge, with the fully-aligned $\mathcal{M}_{\dlmq}$ models achieving a 18.57\% lead over the $\mathcal{M}_{\dbase}$ model on average. The most striking results appear on the highly challenging AIME competition math problems, where the reasoning-pretrained models deliver a 39.32\% improvement over the baseline. This provides conclusive evidence that 
early investment in reasoning data yields compounding returns, becoming the decisive factor in achieving frontier accuracies on the most demanding tasks.

\section{Ablations}
\label{sec:ablations}

\paragraph{Does the scale and diversity  of the reasoning data matter in Pretraining?} 
As detailed in \autoref{tab:PT_gains}, plainly increasing size and diversity of $\datad_{\dres}$ in pretraining has significant improvement on the base model. 
The model pretrained on large, diverse data ($\mathcal{M}_{\dldq}$) achieves an absolute +9.09\% average gain over the model trained on the smaller, less diverse corpus ($\mathcal{M}_{\dshq}$), with the largest gains observed in math, science, and code—domains that explicitly demand structured reasoning.
\ptgpr shows limited sensitivity to diversity due to the nature of tasks that require commonsense and general knowledge. 
In contrast, scaling $\datad_{\dldq}$ with $\datad_{\dshq}$ (high-quality but less diverse) as in $\mathcal{M}_{\dlmq}$ provides minimal further benefit on the reasoning tasks---underscoring that broad exposure to diverse reasoning patterns during pretraining is impactful for building a strong foundation.

\begin{table*}[t]
\begin{center}
\resizebox{0.8\textwidth}{!}{
\begin{tabular}{@{}lccccc@{}}
\toprule
Model &  Average & \sftmath & \sftsci & \sftcode 	& \sftins\\
\midrule
$\mathcal{M}_{\dbase}+\mathrm{SFT}$ & 26.62 & 34.48	& 20.92	& 7.09	& 43.98\\
$\mathcal{M}_{\dbase}+\mathrm{SFT}$(2$\times$) & 34.01 & 48.05 & 40.69 & 14.60 & 32.70\\
\midrule
$\mathcal{M}_{\dshq}+\mathrm{SFT}$ & 37.33	& 50.52	&40.00	&24.76	&34.06\\
$\mathcal{M}_{\dldq}+\mathrm{SFT}$ & 46.70	&60.79&	50.67	&28.57	&46.79\\
$\mathcal{M}_{\dlmq}+\mathrm{SFT}$ & \textbf{50.95}	&64.67	&53.74	&35.55	&49.82\\
\toprule
\end{tabular}
}
\end{center}
\vspace{-1em}
\caption{\textbf{Impact of diverse pretraining $\datad_{\dres}$ on SFT phase.} Doubling SFT for the baseline fails to ``catch up" to reasoning-pretrained models, while the latent advantage of the mixed-quality pretraining ($\mathcal{M}_{\dlmq}$) emerges, making it the top performer.}
\label{tab:Diverse_PT_gains}
\end{table*}

\paragraph{\textit{The Pretraining Advantage Persists and Resists ``Catch-Up" Attempts via SFT.}}
A central question is whether a model without a reasoning-rich pretraining ($\mathcal{M}_{\dbase}$) can compensate for this deficit by undergoing a more intensive SFT phase. We test this ``catch-up" hypothesis by fine-tuning $\mathcal{M}_{\dbase}$ with twice the amount of SFT data. The results in \autoref{tab:Diverse_PT_gains} prove this hypothesis false. While doubling the SFT data improves the baseline's average score by 7.39\%, this enhanced baseline \textbf{still fails to match} the performance of even our weakest reasoning-pretrained model, $\mathcal{M}_{\dshq}+$\textsc{SFT} (+3.32\%). This provides strong evidence that pretraining instills a foundational reasoning capability that cannot be fully replicated by simply scaling the SFT phase.


\paragraph{\textit{Post-SFT, high-quality data reveals latent value.}} The downstream consequences of these pretraining choices become more nuanced after SFT. To isolate and test whether these effects persist into post-training, we finetune all base models with the same high-quality SFT recipe ($\datad_{\dshq}$). Results in \autoref{tab:Diverse_PT_gains} confirm that models pretrained on diverse corpora continue to substantially outperform less diverse counterparts even after SFT, confirming that a diverse pretraining foundation enhances a model's capacity to benefit from SFT. More surprisingly, while the immediate gains of scaling with high-quality but narrow data ($\mathcal{M}_{\dlmq}$) were muted at the pretraining stage, SFT reveals a latent advantage: $\mathcal{M}_{\dlmq}$ achieves an additional $+4.25\%$ improvement over $\mathcal{M}_{\dldq}$ post-SFT. This reveals a critical finding that high-quality but less diverse data may act as a \emph{complementary amplifier}, whose benefits emerge after alignment---underlining the latent impact of quality of data during the pretraining.
\begin{table*}[h!]
\begin{center}
\resizebox{0.8\textwidth}{!}{%
\begin{tabular}{@{}lccccc@{}}
\toprule
Model &  Average &\sftmath & \sftsci & \sftcode & \sftins \\
\midrule
$\mathcal{M}_{\dbase}+\mathrm{SFT}$[$\datad_{\dshq}$] & 29.92 & 42.79 & 35.83	&10.48	&30.59 \\
\midrule
$\mathcal{M}_{\dres}+\mathrm{SFT}$[$\datad_{\dlmq}$] &31.21	&30.91	&27.73	&9.79	&56.41\\
$\mathcal{M}_{\dres}+\mathrm{SFT}$[$\datad_{\dldq}$] & 31.54	&32.28	&28.43	&10.85	&54.61\\
$\mathcal{M}_{\dres}+\mathrm{SFT}$[$\datad_{\dshq}$] & \textbf{44.99}	&58.66	&48.14	&29.63	&43.56\\
\toprule
\end{tabular}
}%
\end{center}
\vspace{-1em}
\caption{\textbf{Impact of diverse SFT $\datad_{\dres}$ on SFT phase.} Fine-tuning on the small, high-quality corpus ($\datad_{\dshq}$) is highly effective, while using large, diverse corpora ($\datad_{\dldq}$) degrades reasoning.}
\label{tab:Diverse_SFT_gains}
\end{table*}


\paragraph{\textit{SFT is dominated by data quality, not diversity.}} 
We finetune all reasoning-pretrained models ($\mathcal{M}_{\dres}$) on each of our distinct reasoning datasets, and report the averaged results in \autoref{tab:Diverse_SFT_gains}. The findings reveal a striking contrast: while diversity is beneficial in pretraining, blindly scaling diverse reasoning data during SFT degrades performance. Models trained with $\datad_{\dldq}$ or $\datad_{\dlmq}$ during SFT underperform relative to those finetuned on the smaller, high-quality, long-CoT dataset, $\datad_{\dshq}$, despite having been exposed to reasoning data during pretraining. In fact, the use of large-scale, mixed-quality data at the SFT stage not only erodes the benefits of reasoning-rich pretraining but can even lead to worse outcomes than the baseline $\mathcal{M}_{\dbase}$ finetuned with $\datad_{\dshq}$ in math, code, and science tasks which benefit from reasoning. This result confirms the widely held view that data quality and long reasoning data is critical for effective SFT \citep{zhou2023lima, zhao2024long, prabhakar2025omnisciencedomainspecializedllmscientific}. Our findings, however, extend this understanding by showing that simply applying high-quality data at every stage is not optimal. Instead, the most effective strategy is \textbf{asymmetric}: pretraining benefits most from broad and diverse reasoning data to establish generalizable priors, whereas \textsc{sft} requires high-quality, reasoning-heavy data for targeted refinement. 

\textbf{\begin{table*}[t]
\begin{center}
\resizebox{0.8\textwidth}{!}{%
\begin{tabular}{@{}lccccc@{}}
\toprule
Model &  Average &\sftmath & \sftsci & \sftcode & \sftins \\
\midrule
$\mathcal{M}_{\dldq}+\mathrm{SFT}$[$\datad_{\dldq}$] & 32.84	&28.38	&35.22	&10.16	&57.61 \\
$\mathcal{M}_{\dldq}+\mathrm{SFT}$[$2\times\datad_{\dldq}$] &32.99	&23.46	&39.65	&11.75	&57.10\\\midrule
$\mathcal{M}_{\dldq}+\mathrm{SFT}$[$\datad_{\dalf}$] & 42.66	&60.95	&47.29	&22.54	&39.87 \\
$\mathcal{M}_{\dldq}+\mathrm{SFT}$[$\datad_{\dalf}^{'}$] &43.04	&61.61	&45.78	& 22.53	&42.23\\
\toprule
\end{tabular}
}%
\end{center}
\caption{\textbf{Impact of scaling reasoning data in SFT phase.} Naively doubling mixed-quality data is detrimental to math reasoning, whereas targeted scaling of high-quality data yields consistent gains.}
\label{tab:scale_SFT_gains}
\end{table*}}

\paragraph{How should we expand reasoning data during SFT?}
We next ablate the effect of scaling reasoning data during the SFT phase by contrasting two strategies: (i) scaling with data of \emph{similar quality and diversity}, and (ii) scaling with data of \emph{higher quality and reasoning depth}.  

As shown in \autoref{tab:scale_SFT_gains}, simply doubling the amount of diverse but mixed-quality data yields negligible improvement in average accuracy with a 4.92\% drop in math accuracy---suggesting that increasing the volume of noisy or shallow reasoning data may dilute the useful signal and actively harm reasoning-specific domains. The small gains in science and code do not offset this regression, highlighting the limits of quantity-driven scaling in SFT.


In contrast, 
when scaling $\datad_{\dalf}$ with high-quality $\datad_{\dshq}$ ($\datad_{\dalf}^{'}$), the average accuracy improves further, with math and instruction-following tasks benefiting most. Importantly, this qualitative expansion is achieved with only a marginal increase in dataset size (0.4\% more samples). These contrasting outcomes provide clear evidence that SFT is a phase of targeted refinement, not broad data absorption; the most effective scaling strategy is to strategically enhance the training corpus with high-quality, reasoning-intensive examples.





\section{Related Work}
\label{sec:related_work}

\paragraph{Reasoning in Pretraining and Midtraining.}  
\citet{cheng-etal-2024-instruction} study \emph{instruction pretraining} by converting raw text into short QA pairs and report gains on general-purpose reasoning tasks that require minimal reasoning. While effective for broad linguistic alignment, their setup does not explicitly target reasoning-intensive domains such as mathematics, graduate level science, or code. Moreover, their pipeline of self-distilled instruction generation demonstrates that Instruct-PT outperforms vanilla PT after instruction tuning, but it does not assess whether these marginal pretraining gains persist once models undergo reasoning-heavy SFT and reinforcement learning. In contrast, our work systematically varies the \emph{complexity, quantity, and diversity} of reasoning-style SFT data—containing intermediate thoughts and answers—across both pretraining and SFT, allowing us to probe whether early exposure yields durable downstream advantages.  

More recent efforts have begun to explore the interplay between pretraining and instruction tuning. \citet{liang2025aligninginstructiontuningpretraining} augment the instruction-tuning pool to better align with the distribution of pretraining data, reinforcing consistency between the two stages. While complementary in spirit, their method is applied only during SFT and does not address whether reasoning-specific supervision at the pretraining stage provides sustained benefits. Similarly, \citet{wang2025octothinker, ai2025rethinkingreflectionpretraining} introduce a \emph{mid-training} phase, continuing pretraining on a small but high-quality reasoning dataset before SFT and RLVR. They report substantial downstream gains, particularly in mathematics benchmarks, highlighting the promise of mid-training interventions. However, because their corpus is heavily math-centric, it is difficult to disentangle whether the improvements stem from scale, complexity, or domain diversity, and the generalizability to science or code remains unclear.  

A complementary direction is pursued by \citet{gandhi2025cognitive}, who inject algorithmically generated “cognitive behavioral” reasoning traces during mid-training, demonstrating improvements after reinforcement learning. This underscores the potential of early reasoning supervision but remains limited in scope: the interventions are restricted to small datasets and narrow tasks, leaving open questions about scalability, diversity, and phase-specific allocation of reasoning data. Our work builds on these insights by conducting the first systematic, large-scale analysis of reasoning data across both pretraining and SFT, providing a principled framework for understanding \emph{when} and \emph{how} reasoning supervision should be applied.

\section{Conclusion}
\label{sec:conclusion}

Our study provides the first systematic investigation of how reasoning data, varying in scale, diversity, and quality, influences {\llm}s across the entire training pipeline. 
We show that reasoning must be introduced early: front-loading into pretraining creates durable foundations that post-training alone cannot recover. 
Crucially, we uncover an asymmetric allocation principle—diversity drives pretraining effectiveness, while quality governs SFT—providing a clear, actionable blueprint for data strategy. Further, we demonstrate that high-quality pretraining data can yield latent benefits activated only during SFT, and that naive SFT scaling with noisy data can be actively harmful. Collectively, these findings challenge the conventional division between pretraining and reasoning, positioning reasoning-aware pretraining as a critical ingredient in building more capable, generalizable, and compute-efficient language models.



\bibliography{iclr2026_conference}
\bibliographystyle{iclr2026_conference}

\appendix

\section{Appendix}

\section{Experiments and Results}
\label{sec:appendix_expts_results}


\sisetup{
  detect-weight = true,
  detect-family = true,
  group-digits  = false,
  round-mode    = places,
  round-precision = 2,
  table-number-alignment = center
}
\newcolumntype{L}{>{\raggedright\arraybackslash}X}

\begin{table}[t]
\centering
\setlength{\tabcolsep}{3.5pt}
\renewcommand{\arraystretch}{1.1}
\footnotesize

\begin{adjustbox}{max width=\linewidth}
\begin{tabularx}{\linewidth}{
  L
  S[table-format=2.2]
  S[table-format=2.2]
  >{\bfseries\columncolor{gray!8}}S[table-format=2.2]
  >{\bfseries\columncolor{gray!8}}S[table-format=2.2]
  >{\bfseries\columncolor{gray!15}}S[table-format=2.2]
}
\toprule
\textbf{Benchmark} &
\multicolumn{1}{c}{\textbf{$\mathcal{M}_{\dbase}$}} &
\multicolumn{1}{c}{\textbf{$\mathcal{M}_{\dshq}$}} &
\multicolumn{1}{>{\columncolor{gray!8}}c}{\textbf{$\mathcal{M}_{\dldq}$}} &
\multicolumn{1}{>{\columncolor{gray!8}}c}{\textbf{$\mathcal{M}_{\dlmq}$}} &
\multicolumn{1}{>{\columncolor{gray!15}}c}{\textbf{$\mathcal{M}_{\dres}$}}\\
\midrule
\textsc{ARC-C}              &  80.89&80.46&81.40&81.83&81.15 \\
\textsc{RACE}             & 73.59 &75.41 & 78.28 & 79.43 &  76.68 \\
\textsc{WinoGrande}               & 70.64 &71.43 & 69.53 & 69.38 &  70.25\\
\textsc{HellaSwag}               & 77.38 & 77.06 & 76.69 & 76.67 &  76.95\\\midrule
\textsc{GSM8K}                & 59.74 & 65.20 & 82.71 & 85.14 & 73.20\\
\textsc{Math-500}           & 34.60 & 40.00 & 68.40 & 59.60 &  50.65 \\\midrule
\textsc{MMLU}            & 61.67 & 61.45 &65.87 & 65.42 &  63.60\\
\textsc{MMLU-Pro}           & 32.59 & 32.34 & 42.89 & 43.56 & 37.85\\\midrule
\textsc{HumanEval}       & 37.44 & 41.04 & 48.63 & 51.68 & 44.70\\
\textsc{HumanEvalPlus}                & 32.59 &  35.03 & 42.74 & 46.28 & 39.16\\
\textsc{Mbpp}          & 41.64 & 47.47 & 48.85 & 51.47 &  47.36\\
\textsc{Mbpp[sanitized]}                & 51.87 & 53.74 & 59.53 & 60.97 &  56.53\\\midrule
\ptmath            & 47.17 & 52.60 & 75.56 & 72.37 &  61.92\\
\ptsci         & 47.13 & 46.90 & 54.38 & 54.49 &  50.72\\
\ptcode    & 40.89 & 44.32 & 49.94 & 52.60 & 46.94 \\
\ptgpr    & 75.63 & 76.09 & 76.48 & 76.83 & 76.25 \\
\midrule
\textbf{Overall}    & 52.70 &  54.98 & {\bfseries 64.09} & 64.07 & {\bfseries 61.05} \\
\bottomrule
\end{tabularx}
\end{adjustbox}

\caption{Breakdown of base model accuracies across benchmarks. With increasing diversity and quality, the difference between $\mathcal{M}_{\dbase}$ and models pretrained with reasoning data increases.}
\label{tab:qwen-main-res}
\vspace{-2.5mm}
\end{table}

\sisetup{
  detect-weight = true,
  detect-family = true,
  group-digits  = false,
  round-mode    = places,
  round-precision = 2,
  table-number-alignment = center
}
\newcolumntype{L}{>{\raggedright\arraybackslash}X}

\begin{table}[t]
\centering
\setlength{\tabcolsep}{3.5pt}
\renewcommand{\arraystretch}{1.1}
\footnotesize

\begin{adjustbox}{max width=\linewidth}
\begin{tabularx}{\linewidth}{
  L
  S[table-format=2.2]
  S[table-format=2.2]
  >{\bfseries\columncolor{gray!8}}S[table-format=2.2]
  >{\bfseries\columncolor{gray!8}}S[table-format=2.2]
  >{\bfseries\columncolor{gray!15}}S[table-format=2.2]
}
\toprule
\textbf{Benchmark} &
\multicolumn{1}{c}{\textbf{$\mathcal{M}_{\dbase}+\mathrm{SFT}$}} &
\multicolumn{1}{c}{\textbf{$\mathcal{M}_{\dshq}+\mathrm{SFT}$}} &
\multicolumn{1}{>{\columncolor{gray!8}}c}{\textbf{$\mathcal{M}_{\dldq}+\mathrm{SFT}$}} &
\multicolumn{1}{>{\columncolor{gray!8}}c}{\textbf{$\mathcal{M}_{\dlmq}+\mathrm{SFT}$}} &
\multicolumn{1}{>{\columncolor{gray!15}}c}{\textbf{$\mathcal{M}_{\dres}+\mathrm{SFT}$}}\\
\midrule
\textsc{IFEval}              &  30.59	&34.06	&46.79	&49.82	&43.56 \\\midrule
\textsc{AIME-24}                & 8.12	& 18.33	& 35.21	& 41.88	& 31.81\\
\textsc{AIME-25}                & 11.88	& 18.12	& 29.38	& 33.12	& 26.87\\
\textsc{GSM8K}                & 81.24	& 86.58	& 91.05	& 92.84	& 90.16\\
\textsc{Math-500}           & 69.9	& 79.05	& 87.50	& 90.85	& 85.80 \\\midrule
\textsc{MMLU}            & 52.14	& 62.9	& 71.15	& 73.49	& 69.18\\
\textsc{MMLU-Pro}           & 39.45	& 48.63	& 53.45	& 55.54	& 52.54\\
\textsc{GPQA-Diamond}           & 15.91	& 8.46	& 27.40	& 32.20	& 22.69\\
\midrule
\textsc{LiveCodeBench}       & 10.48	& 24.76	& 28.57	& 35.55	& 29.63\\\midrule
\sftmath            & 42.79 &	50.52 &	60.79 &	64.67 &	58.66\\
\sftsci         & 35.83 &	40.00 &	50.67 &	53.74 &	48.14\\
\sftcode    & 10.48 &	24.76 &	28.57 &	35.55 &	29.63 \\
\sftins    & 30.59 &	34.06 &	46.79 &	49.82 &	43.56 \\
\midrule
\textbf{Overall}    & 35.52 & 42.32 & 52.28 & 56.14 & {\bfseries 50.25} \\
\bottomrule
\end{tabularx}
\end{adjustbox}

\caption{Breakdown of model accuracies across benchmarks after training SFT phase on the $\mathcal{D}_{\dshq}$. Model pretrained with reasoning data obtains the highest gain after heavy SFT phase of training.}
\label{tab:qwen-main-res}
\vspace{-2.5mm}
\end{table}

\section{Additional Ablations}
\label{sec:data_redun}

\paragraph{Anatomy of high-quality reasoning data in SFT.}
Our previous results establish that SFT benefits immensely from high-quality data, but what precisely constitutes ``quality'' remains unclear. 
In this ablation, we investigate a defining characteristic of such data: the depth and complexity of its reasoning traces. Specifically, we compare datasets that differ both in reasoning length and construction method. The high-quality corpus $\datad_{\dshq}$ consists of answers generated by strong teacher models, characterized by long chain-of-thoughts with an average length exceeding $10$k tokens. In contrast, $\datad_{\dldq}$ provides reasoning data from diverse domains but with much shorter and noisier reasoning traces (average $\sim$550 tokens). This distinction highlights a potential mechanism underlying quality: longer reasoning chains may serve as richer supervisory signals, encouraging models to internalize structured multi-step inference rather than surface-level heuristics.

To test this hypothesis, we extract from $\mathcal{D}_{LLQ}$ only the longest reasoning traces, creating a new dataset $\mathcal{D}_{ALF}$. Although it represents only $\sim$2\% of the original $\mathcal{D}_{LLQ}$ corpus, $\mathcal{D}_{ALF}$ is highly skewed toward domains with inherently deeper reasoning (75\% math, with the remainder in science, code, and general reasoning). We then conduct SFT on top of the $\mathcal{M}_{llq}$ model using both $\mathcal{D}_{LLQ}$ (quantity and diversity) and $\mathcal{D}_{ALF}$ (length-filtered complexity).  

\textbf{\begin{table*}[h!]
\begin{center}
\resizebox{0.8\textwidth}{!}{%
\begin{tabular}{@{}lccccc@{}}
\toprule
Model &  Average &\sftmath & \sftsci & \sftcode 	& \sftins\\\midrule
$\mathcal{M}_\dldq+\mathrm{SFT}$[$\mathcal{D}_{\dldq}$] & 32.84	&28.38	&35.22	&10.16	&57.61 \\
$\mathcal{M}_{\dldq}+\mathrm{SFT}$[$\mathcal{D}_{\dalf}$] &42.71 &	60.95	&47.50	&22.54	&39.87\\
\toprule
\end{tabular}
}%
\end{center}
\caption{\textbf{Impact of depth in reasoning traces in data on SFT phase.} Model trained on longer CoT reasoning data outperforms the one trained on diverse reasoning traces.}
\label{tab:Length_SFT_gains}
\end{table*}}

As shown in \autoref{tab:Length_SFT_gains}, emphasizing depth in reasoning traces has a significant impact on downstream reasoning tasks. While finetuning with $\mathcal{D}_{LLQ}$ yields only modest improvements, switching to the 50 times smaller, filtered by reasoning depth via answer length $\mathcal{D}_{ALF}$ boosts the overall score to 9.87\%, with particularly strong gains in math, science and code. Interestingly, this comes at the cost of slightly reduced accuracy on instruction-following tasks, reflecting a trade-off between breadth and reasoning-specific depth.  These results provide strong evidence that \emph{longer chain-of-thought supervision is a critical marker of quality in \textsc{sft} data}. Even when drawn from a noisy, large-scale corpus, selecting for reasoning depth alone can yield outsized improvements, making length-filtering a simple yet cost-effective heuristic for constructing impactful reasoning datasets for \textsc{sft} phase.


\begin{wrapfigure}{r}{0.6\textwidth}
    \vspace{-6mm}
    \centering    
    \includegraphics[width=0.5\textwidth]{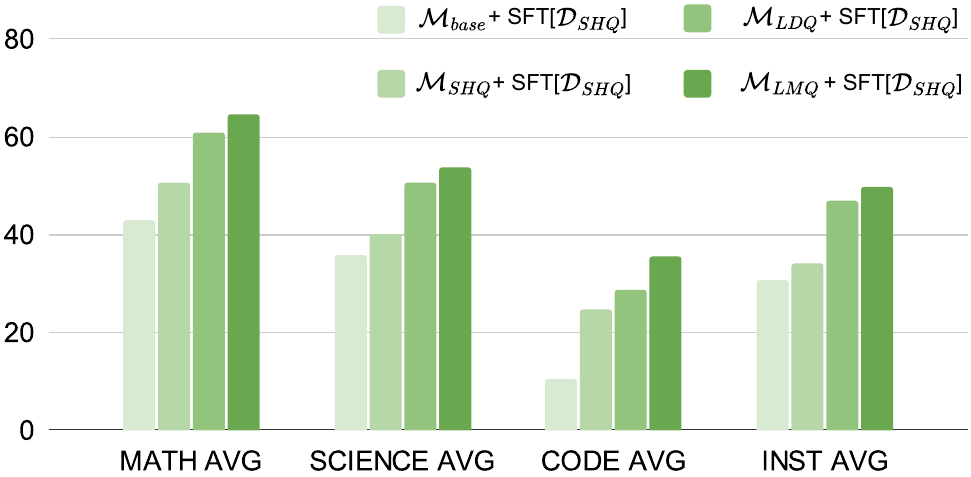}
    \caption{The model that saw the same high-quality data in both pretraining and SFT ($\mathcal{M}_\dshq$) handily beats the baseline ($\mathcal{M}_{base}$) that only saw the data once.}
  \label{fig:overfitting}
  \vspace{-15mm}
\end{wrapfigure}

\paragraph{Data Redundancy Reinforces Foundational Skills, Not Overfitting.}
A critical consideration in our two-phase approach is whether using the same reasoning data in both pretraining and SFT leads to catastrophic forgetting or brittle overfitting, a known concern in sequential fine-tuning \citep{luo2025empiricalstudycatastrophicforgetting,chen2025sftrlearlyinvestigation}. Our results, shown in \autoref{fig:overfitting}, suggest this concern is unfounded and that the opposite is true: for reasoning, strategic redundancy is highly beneficial. The baseline model, $\mathcal{M}_{base}$, exposed to the high-quality $\mathcal{D}_{\dshq}$ data only during SFT, is the lowest performer across all categories. In contrast, $\mathcal{M}_{\dshq}$, which sees this same data in both phases, demonstrates a significant performance uplift, indicating that the second exposure reinforces rather than overwrites learning. We hypothesize this occurs because the two training phases serve different learning functions. During pretraining, the reasoning data is integrated slowly into the model's core representations alongside vast, diverse knowledge, forcing an internalization of abstract logical patterns. The SFT phase then acts not as a new learning task, but as a powerful reinforcement signal on an already-prepared foundation. This benefit is amplified by a diverse pretraining context: the top-performing $\mathcal{M}_{\dlmq}$ model leverages its broad exposure to various reasoning styles to most effectively capitalize on the repeated, high-quality signal from $\mathcal{D}_{\dshq}$. This suggests that data redundancy between pretraining and SFT should be viewed as a powerful mechanism for skill consolidation, where a diverse pretraining builds the capacity for reasoning and redundant SFT sharpens it.


\end{document}